\documentclass[conference]{IEEEtran}
\IEEEoverridecommandlockouts

\usepackage[numbers,sort&compress]{natbib}
\usepackage{amsmath,amssymb,amsfonts}
\usepackage{algorithmic}
\usepackage{graphicx}
\usepackage{textcomp}
\usepackage{xcolor}
\usepackage{multirow}
\usepackage{times}
\usepackage{latexsym}
\usepackage[T1]{fontenc}

\usepackage[utf8]{inputenc}

\usepackage{microtype}

\usepackage{tabularx}
\usepackage{float}

\usepackage{booktabs}

\usepackage{balance}

\usepackage{subcaption} 
\usepackage{caption}

\usepackage{array}

\usepackage{url}

\usepackage{listings}
\usepackage{listingsutf8}
\def\BibTeX{{\rm B\kern-.05em{\sc i\kern-.025em b}\kern-.08em
    T\kern-.1667em\lower.7ex\hbox{E}\kern-.125emX}}

\begin{document}

\title{Adaptation of Embedding Models to Financial Filings via LLM Distillation}

\author{\IEEEauthorblockN{Eliot Brenner, Dominic Seyler, Manjunath Hegde, Andrei Simion, Koustuv Dasgupta, Bing Xiang}
\IEEEauthorblockA{\textit{Goldman Sachs}\\
New York, NY, USA\\
\{Eliot.Brenner, Dominic.Seyler, Manjunath.y.Hegde, Andrei.Simion, Koustuv.x.Dasgupta, Bing.Xiang\} @gs.com}}

\maketitle

\begin{abstract}
    Despite advances in generative large language models (LLMs), practical application of specialized conversational AI
	agents remains constrained by computation costs, latency requirements, and the need for precise domain-specific relevance measures.
	While existing embedding models address the first two constraints, they underperform on information retrieval in specialized
	domains like finance.  This paper introduces a scalable pipeline that trains specialized models from an unlabeled
	corpus using a general purpose retrieval embedding model as foundation.  Our method yields an average of 27.7\% improvement
	in MRR\texttt{@}5, 44.6\% improvement in mean DCG\texttt{@}5 across 14 financial filing types measured over 21,800
	query-document pairs, and improved NDCG on 3 of 4 document classes in FinanceBench.
	We adapt retrieval embeddings (bi-encoder) for RAG, not LLM generators, using LLM-judged relevance to 
	distill domain knowledge into a compact retriever.  There are prior works
	which pair synthetically generated queries with real passages to directly fine-tune the retrieval model.
	Our pipeline differs from these by introducing
	interaction between student and teacher models that interleaves
	retrieval-based mining of hard positive/negative examples from the unlabeled corpus with iterative retraining of
	the student model's weights using these examples.  Each retrieval iteration uses the refined student model
	to mine the corpus for progressively harder training examples for the subsequent training iteration.  
	The methodology provides a cost-effective solution
	to bridging the gap between general-purpose models and specialized domains without requiring labor-intensive
	human annotation.  
\end{abstract}

\begin{IEEEkeywords}
embedding models, domain adaptation, financial text, retrieval-augmented generation, large language models, knowledge distillation.
\end{IEEEkeywords}

\section{Introduction}
Financial services firms are under increasing pressure to extract insights from complex documents like SEC EDGAR filings \cite{sec_edgar}, which pose challenges due to their length, specialized language, and structure \cite{captide_agentic_rag}. The industry's adoption of AI demands solutions that accurately retrieve and contextualize this information efficiently.

Retrieval-Augmented Generation (RAG) is a common approach, particularly for conversational AI in finance. Most RAG systems use pipelines that chunk text and transform these chunks into vector representations for fast semantic search. Embedding quality is key, but often lacking in specialized domains like finance \cite{li2024retrieval,tang2025finmteb}. While Large Language Models (LLMs) understand financial context well, their high cost and performance degradation with long context windows necessitates using them with smaller retrieval embedding models.

Our approach leverages an open-weights LLM to generate training data from SEC filings, enabling the fine-tuning of a smaller, more efficient retrieval embedding model that captures nuanced financial semantics. This model improves relevance for financial queries, reduces computational requirements, and accelerates inference. Our approach collects positive and negative samples for fine-tuning a retrieval model over a much larger corpus than previously possible.

Despite advances in long-context models, production financial systems benefit from smaller, specialized models. Long-context models are computationally prohibitive for high-volume, low-latency tasks and have questionable comprehension abilities \cite{tang2024we}. Our approach provides domain-specific relevance at a lower computational cost, addressing a critical gap in practical applications.

\section{Related Work}
\label{sec:rw}
Recent studies highlight the importance of domain-specific models in finance. \citet{peng2021domain} found that models continuously pretrained on financial text outperform those trained on general text, while \citet{tang2025finmteb} demonstrated performance drops when general embedding models are applied to finance-specific tasks. Furthermore, \citet{li2024retrieval} showed that general models often struggle with specialized vocabulary and document structures in financial filings. \citet{wang2024omnieval} found the choice of retrieval embedding model significantly impacts downstream generator performance in financial contexts. These findings emphasize the need for specialized financial models.

\citet{anderson2024greenback} introduced a closely related work, 
presenting text embeddings fine-tuned on 14.3 million financial query-passage pairs. They demonstrated improved retrieval embedding models via domain-specific training, achieving a 62.8\% Recall@1 versus 39.2\% for OpenAI's best general-purpose embeddings, highlighting the challenges of financial text. Our method differs primarily in its mining of positive examples, using an LLM judge to generate roughly $10^{3}$ positive passages per query (over two iterations), compared to their generally single positive passage per query. We also mine hard negative examples from within the same documents as the positive examples, using LLM relevance judgments and considering all retrieval ranks, whereas they source negatives from fixed ranks within the entire corpus without judging irrelevance. This allows us to make the model better adapted for within-document retrieval. Our approach generates queries by few-shot prompting which encourages the generation of queries which are not as closely tied to particular passages, incorporates document class and query association with a single document class, and enables the calculation of 
IR metrics like mean DCG$@k$ and MRR$@k$ due to LLM-based reward step labeling of \textit{each} of the top-$k$ chunks in each document (i.e., retrieval set).

In the finance domain, \citet{zhao-etal-2024-optimizing} systematically benchmark
RAG pipeline components on a real financial dataset,
underscoring the importance of retrieval quality
and domain-tailored evaluation.

Because SEC reports are highly standardized and long, general RAG
pipelines under-retrieve or duplicate boilerplate.  Concurrent
work on filings-native RAG \citet{Choe2025Hierarchical}
addresses near-duplicate sections in filings at scale,
and DocFinQA \citet{Reddy2024DocFinQA} shows retrieval
is essential on \textit{full filings}.  Our work
complements these by adapting the retriever itself via
LLM-judged distillation to improve within-document passage selection across
14 filing types.

\section{Dataset}
\label{sec:dataset}

\begin{table}[t]
    \caption{Statistics of the Training Corpus}
    \label{tab:filing_summary}
    \centering
    \small
    \setlength{\tabcolsep}{3pt}  
    \begin{tabular}{|l|l|r|r|r|}
        \hline
        \textbf{Category} & \textbf{Type} & \textbf{\shortstack{\# Docs\\(K)}} & \textbf{\shortstack{Avg.\\Chunks}} & \textbf{\shortstack{\# Chunks\\(M)}} \\
        \hline
        \multirow{4}{*}{\shortstack[l]{Financial\\Report}} & 10-K & 67 & 123 & 8.23 \\
        \cline{2-5}
        & 10-Q & 36 & 85 & 3.10 \\
        \cline{2-5}
        & 6-K & 28 & 56 & 1.55 \\
        \cline{2-5}
        & 8-K & 79 & 72 & 5.72 \\
        \hline
        \multirow{3}{*}{Prospectus} & 424B3 & 5 & 440 & 2.55 \\
        \cline{2-5}
        & 424B4 & 0.2 & 1,664 & 0.38 \\
        \cline{2-5}
        & 497K & 12 & 78 & 0.92 \\
        \hline
        \multirow{3}{*}{Amendment} & 485BPOS & 17 & 499 & 8.94 \\
        \cline{2-5}
        & SC 13D/A & 7 & 44 & 0.3 \\
        \cline{2-5}
        & SC 13G/A & 22 & 20 & 0.44 \\
        \hline
        \shortstack[l]{Pricing\\Details} & 424B2 & 101 & 167 & 16.87 \\
        \hline
        \shortstack[l]{Proxy\\Statement} & DEF 14A & 4 & 598 & 2.56 \\
        \hline
        Registration & S-8 & 8 & 24 & 0.18 \\
        \hline
        \shortstack[l]{Ownership\\Report} & SC 13G & 8 & 19 & 0.15 \\
        \hline
        \textbf{Total:} & & \textbf{396} & & \textbf{51.88} \\
        \hline
    \end{tabular}
\end{table}

We collected a dataset by crawling SEC filings documents, amounting to 396,165 documents distributed over 14 filing types, from the SEC website~\cite{sec_edgar}, published in the time frame of January 1st, 2024 to July 31st, 2024.

We divide our dataset as follows: January 1st through June 30th data are used for training and validation purposes, whereas the month of July data are held-out for testing.

Table~\ref{tab:filing_summary} shows the statistics of the training corpus. See \cite{SECEdgarFilerManualVol2} for a description of each filing type.

\section{Method}
\label{sec:method}
\begin{figure}[h]
	\centering
	\includegraphics[width=0.48\textwidth]{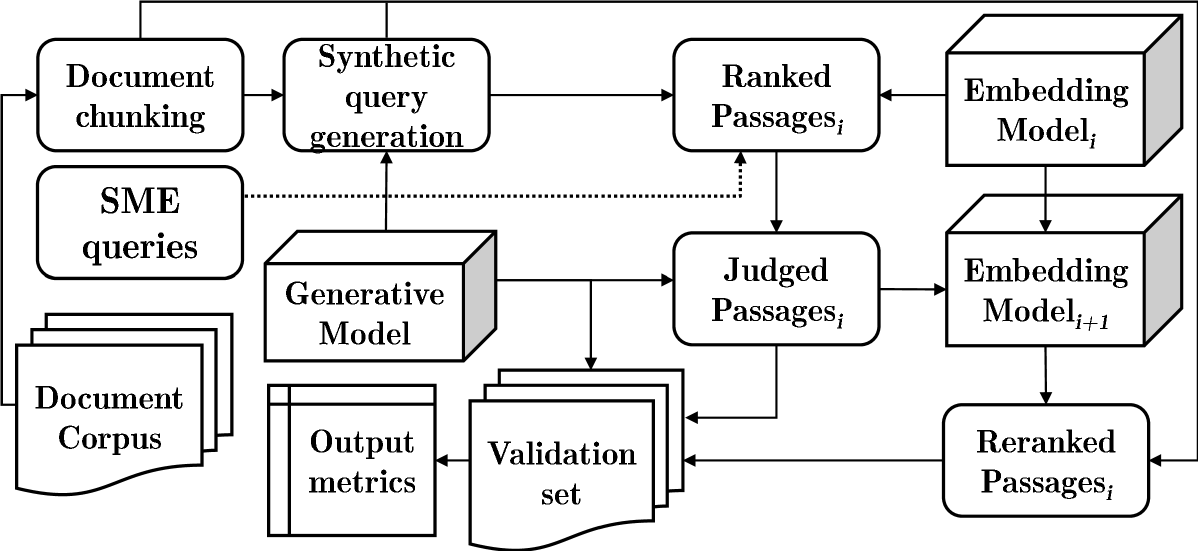}
	\caption{Training/validation pipeline overview. The iterative components of the pipeline are subscripted with ``i''.}
	\label{fig:training-validation-pipeline-overview}
\end{figure}

\begin{figure}[ht]
	\centering
	\includegraphics[width=0.48\textwidth]{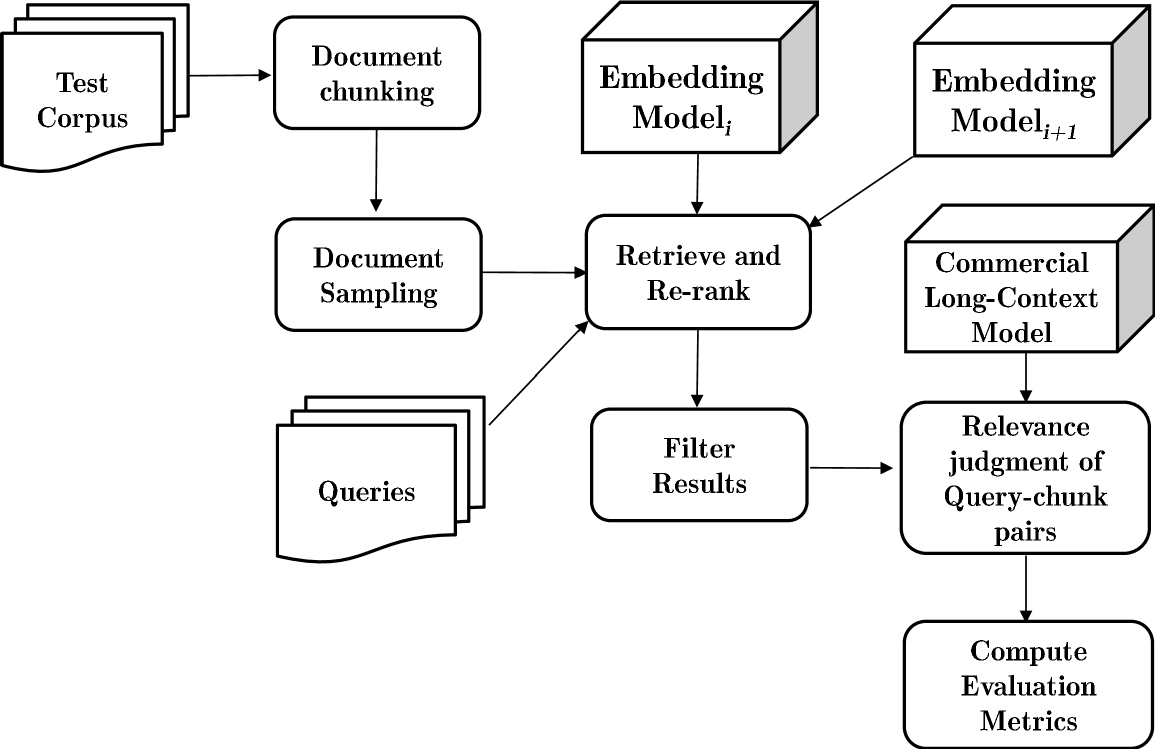}
	\caption{Final evaluation pipeline. The iterative components of the pipeline are subscripted with ``i''. The test corpus is fixed for all iterations.}
	\label{fig:evaluation-pipeline}
\end{figure}

We give a conceptual overview of the main pipeline in Section \ref{subsec:scaffolditerative} (Figure \ref{fig:training-validation-pipeline-overview}
shows the main pipeline and  Figure~\ref{fig:evaluation-pipeline} the evaluation pipeline) and 
then explain certain details and modifications in Sections~\ref{subsec:sparsity}--\ref{subsec:synthetic}.

\subsection{Scaffold Iterative Process}
\label{subsec:scaffolditerative}

\noindent \textbf{Prior to all iterations:}
The corpus $D$ of documents $d\in D$ is divided into the disjoint union of $D_{\rm train-val}$ (for brevity, $D_{\rm tv}$)
and $D_{\rm test}$.
$D_{\rm tv}$ is similarly sub-divided into the disjoint union of $D_{\rm train}$
and $D_{\rm val}$.  Thus, for example
\begin{equation}\label{eqn:Dsplitting}
	D_{\rm tv} = D_{\rm train} \cup D_{\rm val}.
\end{equation}

The corpus is chunked into chunks of between 500 to 1000 characters, respecting
sentence boundaries where possible.  Call the resulting corpus of chunks $C$, and note
that for our corpus, $|C|\approx 5\times 10^{7}$.  The chunk set $C$ inherits the splitting from $D$ into $C_{\rm tv}$, 
$C_{\rm test}$, $C_{\rm train}$ and $C_{\rm val}$.  Denote by $C_d$ the chunks of any fixed $d \in D$.  Thus,
\[
C = C_{\rm tv} \cup C_{\rm test},
\]
\[
C_{\rm tv} = C_{\rm train} \cup C_{\rm val},
\]
\[
C = \bigcup_{d\in D} C_d,
\]
with all the above unions being disjoint.

\noindent \textbf{In the $i$th iteration ($i\geq 0$):} Denote by bi-enc$(i)$ the state of the bi-encoder model at the start of iteration $i$.
Set bi-enc$(0)$ equal to the gte-large model (see Section \ref{sec:model_selection} below), and for $i>0$, set bi-enc$(i)$ equal to the fine-tuned version created in iteration $i-1$.

\noindent \textbf{Step 1.} Provide the teacher LLM with (a sample) of $c\in C_{\rm tv}$ and prompt it to generate queries relevant to each $c$. 
For generating queries using an 
LLM we use methods based on InPars~\cite{bonifacio2022inpars},
which means that we prompt the LLM for a large
number of queries and select the top-scoring queries according to conditional log probability (see
below for more details).  
We compared the ``vanilla'' and ``gbq'' prompt versions 
but opted to use the vanilla version in our work.
For each document we sample 500 chunks and generate one question per chunk. 
Following \cite{bonifacio2022inpars}, we select the top $k=200$ chunks with the highest token probability.  
After this selection, the selected query set is denoted $Q_i$.

\noindent \textbf{Step 2.} For each $q\in Q_i$, create a \textit{sample} $D_q\subset D_{\rm tv}$, inheriting a splitting $D_q=D_{q,t}\cup D_{q,v}$ from the splitting
\eqref{eqn:Dsplitting}.  Define $C_{q,t}:=\cup_{d\in D_{q,t}} C_d$, $C_{q,v}:=\cup_{d\in D_{q,v}} C_d$,
and $C_q:=C_{q,t}\cup C_{q,v}$. (disjoint union)

\noindent \textbf{Step 3.}  Prompt the teacher LLM to assign
an ordinal relevance score $r_{q,c}\in [1,4]\cap \mathbf{Z}$
to a \textit{sample} of the $c\in C_q$. The criteria
provided to the LLM in scoring prompt are given in Table \ref{tab:llm_scoring}
The collection of triples $(q,c,r_{q,c})$ so obtained
serves as a raw training set for the subsequent steps.

\begin{table}[htbp]
	\centering
	\caption{LLM Relevance Scoring Criteria for Passage-Query Matching}
	\label{tab:llm_scoring}
	\resizebox{\columnwidth}{!}{%
	\begin{tabular}{cl}
	\hline
	\textbf{Score} & \textbf{Criteria in LLM Prompt} \\
	\hline
	1 & \begin{tabular}[c]{@{}l@{}}No answer and not relevant - The query is entirely unrelated to the\\ content of the passage and contains no answer.\end{tabular} \\
	\hline
	2 & \begin{tabular}[c]{@{}l@{}}No answer but somewhat relevant - The query has weak relevance to\\ the content of the passage, the connection between them is unclear or\\ incomplete, we can not answer the query based on the passage.\end{tabular} \\
	\hline
	3 & \begin{tabular}[c]{@{}l@{}}Partial answer and moderately relevant - The query is relevant to the\\ content of the answer passage and shows a clear connection, but there\\ may be some gaps or minor inconsistencies in relevance. If we want to\\ answer the query accurately based on the passage as context, the answer\\ will be partial.\end{tabular} \\
	\hline
	4 & \begin{tabular}[c]{@{}l@{}}Explicit answer and perfectly relevant - The passage is perfectly and\\ directly relevant to the query, with a clear and complete connection\\ between them. The answer is not only relevant but also highly accurate,\\ effectively and explicitly answering the query.\end{tabular} \\
	\hline
	\end{tabular}%
	}
	\end{table}

\noindent \textbf{Step 4.}  Extract from the triples $(q,c,r_{q,c})$
contrastive triples $(q,c,c')$ satisfying $r_{q,c}>r_{q,c'}$.
We enforce the following constraints:
\begin{itemize}
	\item We only use positive and negative examples from the same document to form a triple,
	so $d_c=d_{c'}$.
	\item We only consider $(q,c)$ as positive if $r_{q,c}>3$. 
	and negative if $r_{q,c}<3$.
\end{itemize}
After applying these constraints and 
filtering out duplicates we obtain 2.52 million triples 
for training and 950K triples for validation.

Denote the triples so obtained as $(q,c_{\rm rel}, c_{\rm irrel})$,
the notation indicating that $c_{\rm rel}$ is 
strictly more relevant to $q$ than $c_{\rm irrel}$.  We emphasize
that \textit{in all the triples, the queries are synthetic
but the passages are real chunks from the corpus}.

\noindent \textbf{Step 5.  Student model retraining}.  Use the triples
concatenated from Step 4 in iterations $0,\ldots, i$, 
and triplet loss $\mathcal{L}_{\text{triplet}}$ to retrain the student model.
\[
\begin{aligned}
	\mathcal{L}_{\text{triplet}}
&:=
	\max\!\Big(0,\; \alpha + d\!\big({\rm bi-enc}(q), {\rm bi-enc}(c_{\rm rel})\big) 
	\\&
	\quad - d\!\big({\rm bi-enc}(q), {\rm bi-enc}(c_{\rm irrel})\big)\Big)
\end{aligned}
\]
Our encoder model was fine-tuned on the triples dataset discussed prior. 
We choose the following hyperparameters to train the model: Number of training epochs: 2, 
learning rate: $5\times 10^{-7}$, batch size: 128 (16 per GPU), Optimizer: Adam-W.
After training the model triples accuracy on the validation set increased 
from $\sim 95\%$ to $\sim 98\%$ for experimental and the control variant. 
Further, to counteract catastrophic forgetting and stabilize the model's 
performance, we needed to include a large number of triples that were already 
correctly ranked by the base model. 
In addition, the model was highly sensitive to the margin parameter, 
which we found the best setting to be 0.1 (compared to 5, which is the default 
setting in the sentence transformer library).

If the model prior to retraining is denoted bi-enc$(i)$,
then denote the model after retraining by bi-enc$(i+1)$.  

\noindent \textbf{Step 6. Evaluation for Hyperparameter Tuning:}  Using the relevance judgments
$r_{q,c}$ for $c\in C_{q,v}$ and $q\in Q(i)$, calculate
IR metrics (e.g., DCG) for bi-enc$(i)$ and
bi-enc$(i+1)$.

\subsection{Address Sparsity of Answers in Corpus}
\label{subsec:sparsity}
\subsubsection{Associate Document Class to Queries}
Given that each filing type serves a distinct reporting role, the corpus is treated as 14 sub-corpora, for the purposes
of query-generation and search.  To create a unified
embedding model for all filing types, the methodology adapted is as follows:

Each document class (filing type) is associated with specific queries relevant to it. Training data generation (Steps 1-4) and Evaluation on
Validation Set (Step 6) are performed independently for each filing type.  Data from all filing types is mixed (concatenated) only in Step 5, 
bi-encoder model retraining.  Thus, each generated query is linked to a specific filing type.

\subsubsection{Corpus-wide Retrieval to Select $\,C_{q,t}$}
\label{subsubsec:corpuswide}
To address the sparsity of answers to a query $q$ within the documents $D$ of the filing type associated to $q$, in order to form $D_q$
we perform retrieval using bi-enc$(i)$, and $q$ as anchor text, over all of $C_{tv}$.  Set 
\[
\begin{aligned}
C_{\rm tv}(q,K,i) := \; & \mathrm{top-K\;closest} \;c\in C_{t,v}\\
& \mathrm{to} \;q\; \mathrm{under \;bi-enc}(i).
\end{aligned}
\]
Then set
{\small
\[
D_{\rm tv}(q,K,i) := \{ d\in D_{\rm tv} \;|\; C(d)\cap C_{\rm tv}(q,K,i)\neq \emptyset \},
\]
}
\noindent the idea being that defining $D_q$ as $D_{\rm tv}(q,K,i)$ makes it more likely that a higher
proportion of $d\in D_q$ contain an answer to $q$ than documents chosen at random.
Note that as a consequence of these definitions and the characteristics of our corpus (and similar corpora) $|C(D_{\rm tv}(q,K,i))|$ is typically as large as $10^{3}\times K\approx 10^{5}$.
This observation motivates our next set of adjustments to the core method.

\begin{table}
	{ \footnotesize
	\caption{Selected Notation}
	\label{tab:Notation}
	\begin{tabular}{l p{0.68\columnwidth}}
		 \toprule
		 Notation & Definition\\
		 \midrule
		 $D$ & Corpus of documents $d$ \\
		 fold & Generic way of referring to train, validation (val), test, or train-validation (t-v) folds \\
		 $D_{\rm fold}$ & Documents in fold \\ 
		 $C$ & Chunked corpus \\
		 $C_d$ & Chunks of $d\in D$ \\
		 $C_{\rm fold}$ &  $\cup_{d\in D_{\rm fold}}C_d$ \\
		 $Q(i)$ & Queries generated in iteration $i$, $i\geq 0$ \\
		 $D_{q, \rm fold}$ & A sample of $d\in D_{\rm fold}$ most likely to have answers to $q\in Q(i)$ \\
		 $r_{q,c}\in [1,4]\cap \mathbf{Z}$ & Relevance of $c$ to $q$ as assigned by LLM \\
		 $(q,c_{\rm rel}, c_{\rm irrel})$ & Training triple satisfying $r_{q,c_{\rm rel}}> r_{q,, c_{\rm irrel}}$ \\
		 $\mathrm{bi-enc}(i)$ & Student model in iteration $i$, baseline model for $i=0$ and model trained in iteration $i-1$ for $i>0$ \\
		 $\mathrm{metric}_q(i)$ & Metric of $q$ calculated on $c\in C_{q,v}$ using $\mathrm{bi-enc}(i)$ as retriever \\ 
		 $K$ & Positive integer parameter determining the maximum number of $d\in D_{\rm tv}$ considered in-scope for all $q\in Q(i)$ \\
		 $k$ & Positive integer parameter reflecting the number of passages to be used as context in RAG. \\
		 $C_{\rm fold}(q,K,i)$ & Top-$k$ closest $c\in C_{fold}$ to $q\in Q(i)$ under $\mathrm{bi-enc}(i)$ \\
		 $D_{\rm fold}(q,K,i)$ &  $\{ d\in D_{\rm fold} \;|\; C(d)\cap C_{\rm fold}(q,K,i)\neq \emptyset \}$ \\
		 $C_d(q,k,i)$  & $\mathrm{top}-k\;\mathrm{closest}\; c\in C(d)\;\mathrm{under bi-enc}(i)$ \\
		 $C_d(q,k)$ & Union of $C_d(q,k,i)$ and $2k$ randomly sampled $c\in C_d\backslash C_d(q,k,i)$ \\
		 $P$ & Integer parameter such that the top $1/P$ fraction of $(q,d)\in \cup_{q\in Q(i)}\left\{q\right\}\times D_{\rm v}(q,i)$ are evaluated \\
		 $M$ & Integer parameter determining the number of total distinct queries obtained for each document class as $2M$.\\
		 $C_{\rm test}(q,K,i)$ & $\text{top}-K\;\text{closest}\; c\in C_{\rm test} \text{to}\; q\;\text{under bi-enc}(i)$, for $q\in Q$. \\
		 $C_{\rm test}(q,K)$ & $\bigcup_{i=0,\ldots N} C_{\rm test}(q,K,i)$ \\
		 $D_{\rm test}(q,K)$ & $\{d\in D_{\rm test}\; |\; C(d)\cap C_{\rm test}(q,K)\neq \emptyset\}$ \\
		 $C_{\rm test}(q,d,i,k)$ & $\text{top}-k\; c\in C(d)\; \text{closest to}\; q\; \text{under bi-enc}(i)$ \\
		 \bottomrule
		 \end{tabular}
	 }
	 \end{table}

\subsection{Improve Scalability to Large Corpora}
\subsubsection{Random Sampling and Corpus Partition}
\label{subsubsec:randomsampling}
To decrease the computational cost of retrieving $C(D_{\rm tv}(q,K,i))$, at the beginning of iteration $i$ of the pipeline, a sample $D(i)$ is drawn from $D_{\rm tv}$ to serve in place of $D_{\rm tv}$ in iteration $i$.  
To ensure a more equitable distribution over document classes a number of documents is sampled per class, determined so as to achieve $|C(D)|\approx 10^{6}$ per class.
Similarly, to account for the varying distribution of filing type dates over the time window, the train/validation split is performed by choosing a \textit{filing-type dependent} split date in order to achieve a roughly 70\%/30\% training/validation triples split
for each filing type.

\subsubsection{Downsampling within $\,C(D_{\rm tv}(q,K,i))$}
Judging every query-chunk pair would give on the order of $10^{9}$ calls to the LLM, and to cut down on this number, $C(D_{\rm tv}(q,K,i))$ is further downsampled.
We posit that, in spite of any deficiencies in $\mathrm{bi-enc}(i)$, relevance of chunks in $d\in D_{\rm tv}(q,K,i)$ to $q$ still correlates heavily with their similarity to $q$ under $\mathrm{bi-enc}(i)$.  Therefore,
we sample chunks to actually send to the LLM for judgment in the following way:

\textbf{Step 3a}  For each $d\in D_{\rm tv}(q,K,i)$ rank the $c\in C_d$
using distance under bi-enc$(i)$.  Form $C_d(q,i)$ by taking
the union of
\[
\begin{aligned}
C_d(q,k,i):=\; & \mathrm{top}-k\;\mathrm{closest}\;c\in C_d\; \\
        & \text{under bi-enc}(i)
\end{aligned}
\]
and of a random sample of cardinality $2k$ is drawn from
\[
C_d \backslash C_d(q,k,i),
\]
using a parametric, non-uniform probability distribution over the ranks $r=k,\ldots,|C(d)|$
(\textit{ranks} means the ordinals assigned to chunks by scoring each chunk relative to $q$ with $\mathrm{bi-enc}(i)$ as the embedding model).
The probability distribution over ranks is defined as follows: assign
to each rank $r$, the \textit{unnormalized weight}
\[
w_i=\mathrm{exp}^{-\omega (r-k)},\; \omega\;\mathrm{a}\;\mathrm{constant}.
\]
Then, normalize the $w_i$ to sum to 1 and thereby obtain a probability distribution.  In this way, we achieve the aim of making higher ranks
(exponentially) more likely to be sampled than lower ranks.  Note that by definition
\[
|C_d(q,i)|=k+2k=3k << |C_d|.
\] 


For it to be possible to calculate MRR\texttt{@}$k$ and similar metrics at $k$ in Step 6,
we need to have judgments relevance $C_d(q,k,j)$ not only for $j=i$, but also for $j=i+1$.  According
to \textbf{Step 3a} only the former, not the latter, are obtained prior to model fine-tuning.  
Therefore, we have to add an additional step:

\textbf{Step 6a.} For each $(q,c)\in C_d(q,k,i+1)$ (for each $d\in D_{\rm v}(q,i)$ only)
prompt the \textit{judging} generative model to assign relevance score $r(q,c)$.

This enables the calculation of MRR\texttt{@}$k$ and DCG\texttt{@}$k$, but not beyond $k$,
meaning that the hyperparameter $k$ must be chosen with care at initialization, taking into consideration
to the intended use of the retriever.  We choose $k=5$.  Note that prior to computation of all metrics we used
a threshold ($\geq 4$) to binarize the relevance score $r(q,c)$ into $\{0,1\}$.

\subsection{Generation of ``Synthetic'' Queries}
\label{subsec:synthetic}
To address InPars' tendency to generate chunk-specific queries, we generate ``synthetic" queries using a method
which is \textit{not} conditioned on a specific chunk.
Starting with $M$ InPars-generated queries, we augment this set with synthetic queries using few-shot prompting:

\textbf{Step 1a.  Generation of synthetic
	queries via few-shot prompting.}  
Initialize set of synthetic queries as $\emptyset$.  Until there are $M=200$ synthetic queries repeat the following: 
select $5$ InPars queries (``exemplars''); prompt LLM$(i)$
to \textit{generate, as an SME, $5$ additional queries based on the $5$ exemplars}; 
after rejecting any queries which are too short to be questions or fail other simple heuristic quality measures, 
add the generated queries to the synthetic queries set.
Together with the InPars queries this results in $14*400=5600$ total queries.

To encourage query diversity we sample exemplars from distinct clusters, clusters being based on the embedding model bi-enc$(0)$.

\begin{lstlisting}[caption=LLM Question Generation Prompt]
	You are a seasoned financial expert meticulously reviewing earnings call transcripts with a laser focus, and expertly crafting insightful questions to distill the critical insights trends within.
	Your task is to generate a set of queries that financial experts would ask about an earning call transcript.
	
	# Example of queries:
	{list of existing queries}
	
	generate a set of five new queries, you can familiarize yourself with the nature of queries using the data above. each query should be generated in a new line.
	\end{lstlisting}

\subsection{Final Evaluation Pipeline}
\label{subsec:finalevaluationpipeline}
A separate final evaluation pipeline is needed to address biases present when using only the main training/validation pipeline. 
The main pipeline compares bi-enc$(i)$ and bi-enc$(i+1)$ solely on $Q_i$, 
potentially biasing towards bi-enc(i+1). 
The final evaluation pipeline generates new, out-of-sample queries, $Q_{test}$, using the InPars method with sampled passages d from a held-out dataset $D_{test}$. 
This allows for measuring retrieval performance on $D_{test}$ instead of the training/validation set $D_{tv}$. Furthermore, to mitigate biases from the open-weights teacher model, a different 
(commercial) LLM (GPT-4o) is employed as judge in the final evaluation pipeline.

\subsubsection{Scaffolding of Evaluation Pipeline}

\noindent \textbf{Step 1.}  Denote by $N$ the number of iterations of the main
training pipeline.  For each $q\in Q_{\rm test}$, set
\begin{align*}
C_{\rm test}(q,K,i) := \; &\text{top}-K\;\text{closest}\;c\in C_{\rm test}\; \\
&\text{to}\; q\;\text{under bi-enc}(i),
\end{align*}
\[
C_{\rm test}(q,K):=\bigcup_{i=0,\ldots N} C_{\rm test}(q,K,i).
\]
Define
{\small
\[
D_{\rm test}(q,K) := \{d\in D_{\rm test}\; |\; \\ 
C(d)\cap C_{\rm test}(q,K)\neq \emptyset\}.
\]
}

Each of Steps 2 and 3 below is repeated for $i$ ranging over $0,\ldots, N-1$.

\noindent \textbf{Step 2.}  For each $d\in D_{\rm test}(q,K)$, define
\begin{equation}
\label{eqn:C_test}
\begin{split}
C_{\rm test}(q,d,i,k) :=\; & \text{top}-k\; c\in C(d)\; \text{closest to}\; \\
& q\; \text{under bi-enc}(i).
\end{split}
\end{equation}

\noindent \textbf{Step 3.}  For fixed $i\in \{0,\ldots N-1\}$, $j$ is a variable taking values in $\{i, \,i+1\}$.
Based on certain characteristics of $C_{\rm test}(q,d,j,k)$ (to be described below), 
score all $d\in D_{\rm test}(q,K)$ and define $D_{\rm test}^{(i)}(q,K)\subset D_{\rm test}(q,K)$, 
to be the top-scoring $d$, so that $|D_{\rm test}(q,K)|/| D_{\rm test}^{(i)}(q,K) |\approx P$, for $P\in\mathbf{Z}$ fixed hyperparameter.  Define $C_{\rm test}^{(i)}(q,d,i,k)$
by \eqref{eqn:C_test} for $d\in D_{\rm test}^{(i)}$ and as $\emptyset$ for $d\in D_{\rm test} - D_{\rm test}^{(i)}$.  Repeat the following until the confidence intervals of all metrics are small: 
randomly sample $d\in D_{\rm test}^{(i)}(q,K)$; use the LLM to assign $r(q,c)$ to $(q,c)\in C_{\rm test}^{(i)}(q,d,j,k)$; 
recompute metrics and their confidence intervals \texttt{@}$k$ based on all $r(q,c)$ assigned so far.

In lieu of completely specifying the score in \text{Step 3}, we highlight that the following key point: the score depends directly on two factors which can be
computed for any $d\in D_{\rm test}(q,K)$:
\begin{enumerate}
\item The following (modified Hausdorff) \textit{distance}: $\mathrm{dist}\left(C_{\rm test}(q,d,i,k), C_{\rm test}(q,d,i+1,k)\right)$.
\item The following \textit{similarity}, for $S$ a constant: $\mathrm{relu}\left( S - \mathrm{min}_{j=i,i+1}
\mathrm{dist}\left( q,C_{\rm test}(q,d,j,1) \right)\right)$.
\end{enumerate}
These factors embody the propensity of $d$ to contain text relevant to $q$ and the
supervised fine tuning in iteration $i$ to change the ranking of $C(d)$ in a meaningful manner.
The stopping criterion for the evaluation in \textbf{Step 3} which we used is that, for each metric, the standard error is $<5\%$ of the estimated mean.
The variation in $D_{\rm test}^{(i)}$ and of the $(q,c)$ actually evaluated, explains why in 
 Table \ref{tab:performance-metrics-complete-iter1}, we have to report ``counts'' of $(q,c)$ underlying the reported
 metrics and also why the experiment metrics for iteration $0$ do not match the base metrics for iteration $1$, 
 although the model underlying these metrics is the same.

 \section{Model selection}
 \label{sec:model_selection}
 As our embedding model of choice we use the large variant of the 
 General Text Embeddings (GTE) model~\cite{li2023towards}. 
 It was chosen due to it being a widely-used model optimized for 
 information retrieval downstream tasks. Further, its size of 335 million 
 parameters makes it highly scalable and easy to use in a production setting. 
 We prefer GTE-large over the smaller MiniLM model~\cite{wang2020minilm} 
 due to GTE's superior performance~\cite{li2023towards}. 
 As our generative LLM we chose the Llama-3.1-70B-Instruct 
 \cite{meta-llama-llama-3.1-70b-instruct} as it presented the 
 best performance among open-weight models of that size~\cite{artificialanalysis2024}. 
 We chose the 70B parameter version of the model, 
 as we could not achieve an adequately large throughput on our hardware 
 (8x Nvidia H100) using a larger variant of the model.

\section{Results}
\subsection{Quantitative Results}

\begin{figure}[h] 
	\centering
	\begin{subfigure}{0.48\columnwidth}
	\centering
	\includegraphics[width=\textwidth]{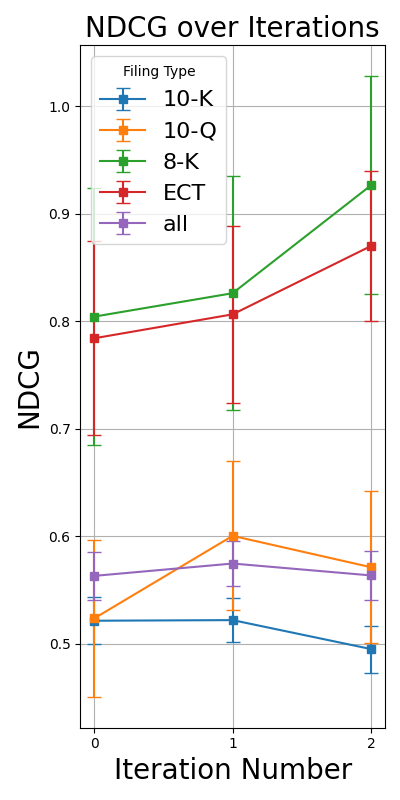}
	\end{subfigure}%
	\hfill
	\begin{subfigure}{0.48\columnwidth}
	\centering
	\includegraphics[width=\textwidth]{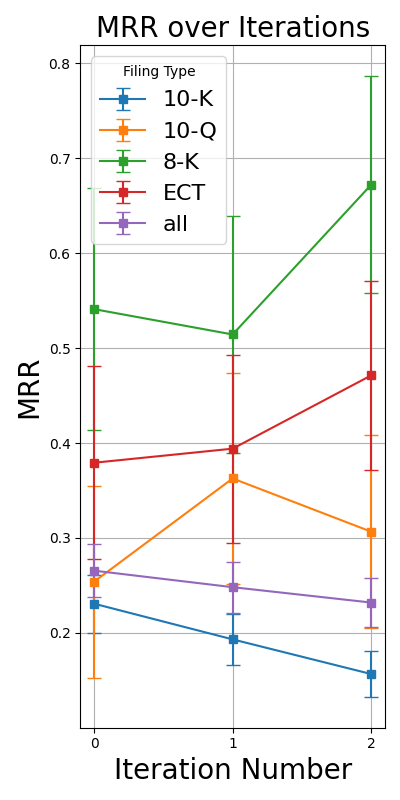}
	\end{subfigure}
	\caption{Metrics with standard errors as measured on FinanceBench Evaluation Set}
	\label{fig:FinanceBench}
\end{figure}

Table~\ref{tab:performance-metrics-complete-iter1} reports performance metrics, calculated as described in Section \ref{subsec:finalevaluationpipeline} for different filing types across two training iterations. 
Table~\ref{tab:FinanceBench} and Figure~\ref{fig:FinanceBench} report performance metrics of bi-enc$(i)$ for $i=0,1,2$ on the public evaluation set of FinanceBench \cite{islam2023financebench}.  Although we did not have access to the FinanceBench training set, we performed this ``out-of-distribution'' evaluation
to enhance reproducibility and provide a stringent challenge for our methods.
We (deterministically) obtained relevance labels $r(q,c)$ for \text{all} $c\in C(d)$
using the ``evidence passages'' present in FinanceBench, as explained below in Section \ref{subsec:qualitative}.
In all reporting tables, the metric followed by ``d'' indicates the Cohen's $d$ \cite{cohen1988statistical}, a widely used statistical measure of effect size.
The Cohen's $d$ values indicate that the training of the base embedding model in the first iteration made a statistically significant difference in ranking performance. 
In the second training iteration, although most metrics remain positive or show improvement in the measured values compared to their baselines, the changes are smaller.
\begin{table*}
	\centering
	\caption{Result for final evaluation pipeline on all filing types. 
	In ``Iteration $i$'', the ``Base'', resp. ``Exp'' 
	version of the metric$@5$ is calculated with bi-enc$i$ (resp. bi-enc$(i+1)$) as the retriever. 
	$\mu$ denotes the average.}
	\begin{tabular}{|l|l|l|l|l|l|l|l|l|}
		\hline
		 & \multicolumn{4}{c|}{\textbf{Iteration 0}} & \multicolumn{4}{c|}{\textbf{Iteration 1}} \\
		\hline
		Type & Count & Base & Exp & Cohen's d & Count & Base & Exp & Cohen's d\\
		\hline
		485BPOS & 1430 & \raisebox{-0.5\height}{\begin{tabular}[c]{@{}l@{}}MRR: \(0.14\)\\ $\mu$DCG: \(0.2\)\end{tabular}} & \raisebox{-0.5\height}{\begin{tabular}[c]{@{}l@{}}MRR: \(0.18\)\\ $\mu$DCG: \(0.27\)\end{tabular}} & \raisebox{-0.5\height}{\begin{tabular}[c]{@{}l@{}}MRR: \(0.099\)\\ $\mu$DCG: \(0.13\)\end{tabular}} & 1430 & \raisebox{-0.5\height}{\begin{tabular}[c]{@{}l@{}}MRR: \(0.18\)\\ $\mu$DCG: \(0.27\)\end{tabular}} & \raisebox{-0.5\height}{\begin{tabular}[c]{@{}l@{}}MRR: \(0.18\)\\ $\mu$DCG: \(0.27\)\end{tabular}} & \raisebox{-0.5\height}{\begin{tabular}[c]{@{}l@{}}MRR: \(0.0048\)\\ $\mu$DCG: \(0.005\)\end{tabular}}\\
		\hline
		6-K & 1801 & \raisebox{-0.5\height}{\begin{tabular}[c]{@{}l@{}}MRR: \(0.11\)\\ $\mu$DCG: \(0.14\)\end{tabular}} & \raisebox{-0.5\height}{\begin{tabular}[c]{@{}l@{}}MRR: \(0.16\)\\ $\mu$DCG: \(0.21\)\end{tabular}} & \raisebox{-0.5\height}{\begin{tabular}[c]{@{}l@{}}MRR: \(0.15\)\\ $\mu$DCG: \(0.17\)\end{tabular}} & 1801 & \raisebox{-0.5\height}{\begin{tabular}[c]{@{}l@{}}MRR: \(0.18\)\\ $\mu$DCG: \(0.24\)\end{tabular}} & \raisebox{-0.5\height}{\begin{tabular}[c]{@{}l@{}}MRR: \(0.19\)\\ $\mu$DCG: \(0.25\)\end{tabular}} & \raisebox{-0.5\height}{\begin{tabular}[c]{@{}l@{}}MRR: \(0.015\)\\ $\mu$DCG: \(0.025\)\end{tabular}}\\
		\hline
		SC 13D/A & 1799 & \raisebox{-0.5\height}{\begin{tabular}[c]{@{}l@{}}MRR: \(0.091\)\\ $\mu$DCG: \(0.13\)\end{tabular}} & \raisebox{-0.5\height}{\begin{tabular}[c]{@{}l@{}}MRR: \(0.13\)\\ $\mu$DCG: \(0.21\)\end{tabular}} & \raisebox{-0.5\height}{\begin{tabular}[c]{@{}l@{}}MRR: \(0.14\)\\ $\mu$DCG: \(0.17\)\end{tabular}} & 1381 & \raisebox{-0.5\height}{\begin{tabular}[c]{@{}l@{}}MRR: \(0.14\)\\ $\mu$DCG: \(0.2\)\end{tabular}} & \raisebox{-0.5\height}{\begin{tabular}[c]{@{}l@{}}MRR: \(0.13\)\\ $\mu$DCG: \(0.2\)\end{tabular}} & \raisebox{-0.5\height}{\begin{tabular}[c]{@{}l@{}}MRR: \(-0.0042\)\\ $\mu$DCG: \(0.0077\)\end{tabular}}\\
		\hline
		8-K & 2460 & \raisebox{-0.5\height}{\begin{tabular}[c]{@{}l@{}}MRR: \(0.13\)\\ $\mu$DCG: \(0.16\)\end{tabular}} & \raisebox{-0.5\height}{\begin{tabular}[c]{@{}l@{}}MRR: \(0.16\)\\ $\mu$DCG: \(0.21\)\end{tabular}} & \raisebox{-0.5\height}{\begin{tabular}[c]{@{}l@{}}MRR: \(0.09\)\\ $\mu$DCG: \(0.12\)\end{tabular}} & 1858 & \raisebox{-0.5\height}{\begin{tabular}[c]{@{}l@{}}MRR: \(0.17\)\\ $\mu$DCG: \(0.22\)\end{tabular}} & \raisebox{-0.5\height}{\begin{tabular}[c]{@{}l@{}}MRR: \(0.18\)\\ $\mu$DCG: \(0.23\)\end{tabular}} & \raisebox{-0.5\height}{\begin{tabular}[c]{@{}l@{}}MRR: \(0.022\)\\ $\mu$DCG: \(0.029\)\end{tabular}}\\
		\hline
		424B4 & 369 & \raisebox{-0.5\height}{\begin{tabular}[c]{@{}l@{}}MRR: \(0.1\)\\ $\mu$DCG: \(0.14\)\end{tabular}} & \raisebox{-0.5\height}{\begin{tabular}[c]{@{}l@{}}MRR: \(0.14\)\\ $\mu$DCG: \(0.22\)\end{tabular}} & \raisebox{-0.5\height}{\begin{tabular}[c]{@{}l@{}}MRR: \(0.13\)\\ $\mu$DCG: \(0.16\)\end{tabular}} & 369 & \raisebox{-0.5\height}{\begin{tabular}[c]{@{}l@{}}MRR: \(0.15\)\\ $\mu$DCG: \(0.24\)\end{tabular}} & \raisebox{-0.5\height}{\begin{tabular}[c]{@{}l@{}}MRR: \(0.16\)\\ $\mu$DCG: \(0.25\)\end{tabular}} & \raisebox{-0.5\height}{\begin{tabular}[c]{@{}l@{}}MRR: \(0.0087\)\\ $\mu$DCG: \(0.028\)\end{tabular}}\\
		\hline
		SC 13G/A & 1719 & \raisebox{-0.5\height}{\begin{tabular}[c]{@{}l@{}}MRR: \(0.17\)\\ $\mu$DCG: \(0.29\)\end{tabular}} & \raisebox{-0.5\height}{\begin{tabular}[c]{@{}l@{}}MRR: \(0.27\)\\ $\mu$DCG: \(0.56\)\end{tabular}} & \raisebox{-0.5\height}{\begin{tabular}[c]{@{}l@{}}MRR: \(0.26\)\\ $\mu$DCG: \(0.34\)\end{tabular}} & 1013 & \raisebox{-0.5\height}{\begin{tabular}[c]{@{}l@{}}MRR: \(0.32\)\\ $\mu$DCG: \(0.6\)\end{tabular}} & \raisebox{-0.5\height}{\begin{tabular}[c]{@{}l@{}}MRR: \(0.32\)\\ $\mu$DCG: \(0.61\)\end{tabular}} & \raisebox{-0.5\height}{\begin{tabular}[c]{@{}l@{}}MRR: \(-0.012\)\\ $\mu$DCG: \(0.00083\)\end{tabular}}\\
        \hline
        DEF 14A & 1249 & \raisebox{-0.5\height}{\begin{tabular}[c]{@{}l@{}}MRR: \(0.23\)\\ $\mu$DCG: \(0.27\)\end{tabular}} & \raisebox{-0.5\height}{\begin{tabular}[c]{@{}l@{}}MRR: \(0.3\)\\ $\mu$DCG: \(0.41\)\end{tabular}} & \raisebox{-0.5\height}{\begin{tabular}[c]{@{}l@{}}MRR: \(0.17\)\\ $\mu$DCG: \(0.25\)\end{tabular}} & 769 & \raisebox{-0.5\height}{\begin{tabular}[c]{@{}l@{}}MRR: \(0.36\)\\ $\mu$DCG: \(0.48\)\end{tabular}} & \raisebox{-0.5\height}{\begin{tabular}[c]{@{}l@{}}MRR: \(0.36\)\\ $\mu$DCG: \(0.5\)\end{tabular}} & \raisebox{-0.5\height}{\begin{tabular}[c]{@{}l@{}}MRR: \(-0.002\)\\ $\mu$DCG: \(0.017\)\end{tabular}}\\
        \hline
        424B3 & 1108 & \raisebox{-0.5\height}{\begin{tabular}[c]{@{}l@{}}MRR: \(0.11\)\\ $\mu$DCG: \(0.16\)\end{tabular}} & \raisebox{-0.5\height}{\begin{tabular}[c]{@{}l@{}}MRR: \(0.13\)\\ $\mu$DCG: \(0.22\)\end{tabular}} & \raisebox{-0.5\height}{\begin{tabular}[c]{@{}l@{}}MRR: \(0.072\)\\ $\mu$DCG: \(0.12\)\end{tabular}} & 1108 & \raisebox{-0.5\height}{\begin{tabular}[c]{@{}l@{}}MRR: \(0.16\)\\ $\mu$DCG: \(0.26\)\end{tabular}} & \raisebox{-0.5\height}{\begin{tabular}[c]{@{}l@{}}MRR: \(0.16\)\\ $\mu$DCG: \(0.25\)\end{tabular}} & \raisebox{-0.5\height}{\begin{tabular}[c]{@{}l@{}}MRR: \(-0.012\)\\ $\mu$DCG: \(-0.0064\)\end{tabular}}\\
        \hline
        10-K & 1349 & \raisebox{-0.5\height}{\begin{tabular}[c]{@{}l@{}}MRR: \(0.13\)\\ $\mu$DCG: \(0.16\)\end{tabular}} & \raisebox{-0.5\height}{\begin{tabular}[c]{@{}l@{}}MRR: \(0.15\)\\ $\mu$DCG: \(0.21\)\end{tabular}} & \raisebox{-0.5\height}{\begin{tabular}[c]{@{}l@{}}MRR: \(0.063\)\\ $\mu$DCG: \(0.11\)\end{tabular}} & 1349 & \raisebox{-0.5\height}{\begin{tabular}[c]{@{}l@{}}MRR: \(0.18\)\\ $\mu$DCG: \(0.25\)\end{tabular}} & \raisebox{-0.5\height}{\begin{tabular}[c]{@{}l@{}}MRR: \(0.19\)\\ $\mu$DCG: \(0.26\)\end{tabular}} & \raisebox{-0.5\height}{\begin{tabular}[c]{@{}l@{}}MRR: \(0.015\)\\ $\mu$DCG: \(0.025\)\end{tabular}}\\
        \hline
        497K & 1759 & \raisebox{-0.5\height}{\begin{tabular}[c]{@{}l@{}}MRR: \(0.19\)\\ $\mu$DCG: \(0.23\)\end{tabular}} & \raisebox{-0.5\height}{\begin{tabular}[c]{@{}l@{}}MRR: \(0.2\)\\ $\mu$DCG: \(0.25\)\end{tabular}} & \raisebox{-0.5\height}{\begin{tabular}[c]{@{}l@{}}MRR: \(0.029\)\\ $\mu$DCG: \(0.047\)\end{tabular}} & 1467 & \raisebox{-0.5\height}{\begin{tabular}[c]{@{}l@{}}MRR: \(0.23\)\\ $\mu$DCG: \(0.29\)\end{tabular}} & \raisebox{-0.5\height}{\begin{tabular}[c]{@{}l@{}}MRR: \(0.23\)\\ $\mu$DCG: \(0.3\)\end{tabular}} & \raisebox{-0.5\height}{\begin{tabular}[c]{@{}l@{}}MRR: \(0.0079\)\\ $\mu$DCG: \(0.021\)\end{tabular}}\\
        \hline
        424B2 & 1214 & \raisebox{-0.5\height}{\begin{tabular}[c]{@{}l@{}}MRR: \(0.25\)\\ $\mu$DCG: \(0.36\)\end{tabular}} & \raisebox{-0.5\height}{\begin{tabular}[c]{@{}l@{}}MRR: \(0.31\)\\ $\mu$DCG: \(0.47\)\end{tabular}} & \raisebox{-0.5\height}{\begin{tabular}[c]{@{}l@{}}MRR: \(0.12\)\\ $\mu$DCG: \(0.16\)\end{tabular}} & 1008 & \raisebox{-0.5\height}{\begin{tabular}[c]{@{}l@{}}MRR: \(0.31\)\\ $\mu$DCG: \(0.47\)\end{tabular}} & \raisebox{-0.5\height}{\begin{tabular}[c]{@{}l@{}}MRR: \(0.31\)\\ $\mu$DCG: \(0.47\)\end{tabular}} & \raisebox{-0.5\height}{\begin{tabular}[c]{@{}l@{}}MRR: \(-0.0063\)\\ $\mu$DCG: \(0.0069\)\end{tabular}}\\
        \hline
        10-Q & 2497 & \raisebox{-0.5\height}{\begin{tabular}[c]{@{}l@{}}MRR: \(0.12\)\\ $\mu$DCG: \(0.15\)\end{tabular}} & \raisebox{-0.5\height}{\begin{tabular}[c]{@{}l@{}}MRR: \(0.16\)\\ $\mu$DCG: \(0.23\)\end{tabular}} & \raisebox{-0.5\height}{\begin{tabular}[c]{@{}l@{}}MRR: \(0.11\)\\ $\mu$DCG: \(0.16\)\end{tabular}} & 1416 & \raisebox{-0.5\height}{\begin{tabular}[c]{@{}l@{}}MRR: \(0.23\)\\ $\mu$DCG: \(0.31\)\end{tabular}} & \raisebox{-0.5\height}{\begin{tabular}[c]{@{}l@{}}MRR: \(0.24\)\\ $\mu$DCG: \(0.34\)\end{tabular}} & \raisebox{-0.5\height}{\begin{tabular}[c]{@{}l@{}}MRR: \(0.028\)\\ $\mu$DCG: \(0.05\)\end{tabular}}\\
        \hline
        SC 13G & 1601 & \raisebox{-0.5\height}{\begin{tabular}[c]{@{}l@{}}MRR: \(0.2\)\\ $\mu$DCG: \(0.32\)\end{tabular}} & \raisebox{-0.5\height}{\begin{tabular}[c]{@{}l@{}}MRR: \(0.25\)\\ $\mu$DCG: \(0.45\)\end{tabular}} & \raisebox{-0.5\height}{\begin{tabular}[c]{@{}l@{}}MRR: \(0.13\)\\ $\mu$DCG: \(0.18\)\end{tabular}} & 1395 & \raisebox{-0.5\height}{\begin{tabular}[c]{@{}l@{}}MRR: \(0.26\)\\ $\mu$DCG: \(0.42\)\end{tabular}} & \raisebox{-0.5\height}{\begin{tabular}[c]{@{}l@{}}MRR: \(0.25\)\\ $\mu$DCG: \(0.42\)\end{tabular}} & \raisebox{-0.5\height}{\begin{tabular}[c]{@{}l@{}}MRR: \(-0.017\)\\ $\mu$DCG: \(-0.0084\)\end{tabular}}\\
        \hline
        S-8 & 1482 & \raisebox{-0.5\height}{\begin{tabular}[c]{@{}l@{}}MRR: \(0.17\)\\ $\mu$DCG: \(0.21\)\end{tabular}} & \raisebox{-0.5\height}{\begin{tabular}[c]{@{}l@{}}MRR: \(0.2\)\\ $\mu$DCG: \(0.26\)\end{tabular}} & \raisebox{-0.5\height}{\begin{tabular}[c]{@{}l@{}}MRR: \(0.085\)\\ $\mu$DCG: \(0.12\)\end{tabular}} & 1482 & \raisebox{-0.5\height}{\begin{tabular}[c]{@{}l@{}}MRR: \(0.22\)\\ $\mu$DCG: \(0.26\)\end{tabular}} & \raisebox{-0.5\height}{\begin{tabular}[c]{@{}l@{}}MRR: \(0.22\)\\ $\mu$DCG: \(0.26\)\end{tabular}} & \raisebox{-0.5\height}{\begin{tabular}[c]{@{}l@{}}MRR: \(-5.6\times 10^{-5}\)\\ $\mu$DCG: \(0.013\)\end{tabular}}\\
        \hline
    \end{tabular}
    \label{tab:performance-metrics-complete-iter1}
    \strut\
\end{table*}

\begin{table*}[htbp]
\centering
\caption{Results on FinanceBench: comparing \textit{overall, not $@k$} metric values of gte-large (``Base'') to bi-enc$(1)$ (``Exp''). }
\begin{tabular}{|l|l|l|l|l|l|l|l|}
		\hline
		Type & Count & MRR Base & MRR Exp & MRR d & $\mu$NDCG Base & $\mu$NDCG Exp & $\mu$NDCG d \\
		\hline
		10-K & 112 & 0.23 $\pm$ 0.03 & 0.19 $\pm$ 0.03 & -1.30 & 0.52 $\pm$ 0.02 & 0.52 $\pm$ 0.02 & 0.027 \\
		10-Q & 15 & 0.25 $\pm$ 0.1 & 0.36 $\pm$ 0.1 & 1.00 & 0.52 $\pm$ 0.07 & 0.6 $\pm$ 0.07 & 1.100 \\
		8-K & 9 & 0.54 $\pm$ 0.1 & 0.51 $\pm$ 0.1 & -0.21 & 0.8 $\pm$ 0.1 & 0.83 $\pm$ 0.1 & 0.190 \\
		ECT & 14 & 0.38 $\pm$ 0.1 & 0.39 $\pm$ 0.1 & 0.15 & 0.78 $\pm$ 0.09 & 0.81 $\pm$ 0.08 & 0.260 \\
		\textbf{all} & \textbf{150} & \textbf{0.27 $\pm$ 0.03} & \textbf{0.25 $\pm$ 0.03} & \textbf{-0.63} & \textbf{0.56 $\pm$ 0.02} & \textbf{0.57 $\pm$ 0.02} & \textbf{0.530} \\
		\hline
\end{tabular}
\label{tab:FinanceBench}
\end{table*}

\begin{table*}
	\centering
	\caption{List of Example Passages with Improvements in Experimental Model on Filings and Finance Bench Corpora}
	\label{tab:example-passages-filing-corpus}
	{%
	  \renewcommand{\arraystretch}{1.3}%
	  \setlength{\extrarowheight}{0pt}%
	  \begin{tabular}{| >{\raggedright\arraybackslash}p{0.2\textwidth} | >{\raggedright\arraybackslash}p{0.3\textwidth} | >{\raggedright\arraybackslash}p{0.3\textwidth} | >{\raggedright\arraybackslash}p{0.1\textwidth} |}
		\hline
		Pattern Name & Query & Promoted Passage & Class \\
		\hline
		\multirow[t]{3}{*}{technical exact match}
		  & What is the par value of the company's common stock?
		  & In connection with the Reverse Stock Split, there was no change in the par value per share of \$0.001.
		  & \multirow[t]{2}{*}{10-K} \\
		\cline{2-3}
		  & What is the interest rate range that the company will pay on its indebtedness under the Revolving Credit Facility for the current period?
		  & .. and any outstanding loans under the Revolving Facility will bear interest at either an Adjusted Term SOFR plus a margin of 1.00\% to 1.75\% or an Adjusted Base Rate plus a margin of 0\% to 0.75\%.
		  & \\
		\cline{2-4}
		  & Which region had the worst topline performance for MGM during FY2022?
		  & Las Vegas Strip Resorts Net revenues of \$8.4 billion in the current year compared to \$4.7 billion in the prior year, an increase of 77\% ...
		  & \raisebox{-0.5\height}{\begin{tabular}[c]{@{}l@{}}ECT\\(FinanceBench)\end{tabular}} \\
		\hline
		\multirow[t]{2}{*}{semantic understanding}
		  & What steps is the company taking to mitigate the risk of not being able to raise the necessary capital to execute its business strategy?
		  & Management plans to address the concerns, as needed, by (a) utilizing recent financing obtained through notes payable; (b) utilizing current lines of credit
		  & 10-K \\
		\cline{2-4}
		  & What potential risks and uncertainties does the company's CEO believe could impact the company's business due to current market conditions and geopolitical conflicts?
		  & Furthermore, the capital and credit markets may be adversely affected by regional conflicts around the world and the possibility of a wider global conflict, global sanctions imposed in response to regional conflicts or an energy crisis.
		  & 10-Q \\
		\hline
		\multirow[t]{3}{*}{table retrieval}
		  & What is the change in the company's CEO's total compensation from the previous year?
		  & For more information, please refer to the “Compensation Discussion and Analysis,” as well as the “Narrative Disclosure to Summary Compensation Table and Grants of Plan-Based Awards Table.”Name and Principal PositionFiscal YearSalary (\$)Bonus (\$)Stock Awards (\$)(1)Non-equity Incentive Plan Compensation (\$)(2)All Other Compensation (\$)(3)Total (\$)Johanna 'Hanneke' Faber(4)Chief Executive Officer2024422,075 2,679,676 2,920,689 675,000 320,306 7,017,746...
		  & DEF-14A \\
		\cline{2-4}
		  & How much was the Real change in Sales for AMCOR in FY 2023 vs FY 2022, if we exclude the impact of FX movement, passthrough costs and one-off items?
		  & Net income attributable to Amcor 109 109 109 7.3 181 181 181 12.3 Net income attributable to non-controlling interests 3 3 4 4 Tax expense 103 103 68 68 Interest expense, net 35 35 70 70 Depreciation and amortization 145 144 EBITDA, EBIT, Net income and EPS 395 250 109 7.3 467 323 181 12.3 ...
		  & \raisebox{-0.5\height}{\begin{tabular}[c]{@{}l@{}}ECT\\(Financebench)\end{tabular}} \\
		\cline{2-4}
		  & Does 3M have a reasonably healthy liquidity profile based on its quick ratio for Q2 of FY2023? If the quick ratio is not relevant to measure liquidity, please state that and explain why.
		  & Inventories Finished goods 2,526 2,497 Work in process 1,527 1,606 Raw materials and supplies 1,227 1,269 Total inventories 5,280 5,372 Prepaids 674 435 Other current assets 539 456 Total current assets 15,754 14,688 Current liabilities Short-term borrowings and current portion of long-term debt \$ 3,033 \$ 1,938 Accounts payable 3,231 3,183 Accrued payroll...
		  & \raisebox{-0.5\height}{\begin{tabular}[c]{@{}l@{}}10-Q\\(Financebench)\end{tabular}} \\
		\hline
	  \end{tabular}%
	}
  \end{table*}

  \begin{table*}[!t]
	\centering
	\caption{List of Example Passages Showing Negative Experiment Results on FinanceBench 10-K Subcorpus}
	\label{tab:example-passages-financebench-negatives}
	\begin{tabular}{|>{\raggedright\arraybackslash}p{0.2\textwidth}|>{\raggedright\arraybackslash}p{0.35\textwidth}|>{\raggedright\arraybackslash}p{0.35\textwidth}|}
		\hline
		Pattern Name & Query & Demoted Passage \\
		\hline
		Multi-faceted Question & Among operations, investing, and financing activities, which brought in the most (or lost the least) cash flow for AMD in FY22? &  assets (1,197) (920) (231) Payables to related parties 379 7 (135) Accounts payable 931 801 (513) Accrued liabilities and other 546 526 574 Net cash provided by operating activities 3,565 3,521 1,071 Cash flows from investing activities: Purchases of property and equipment (450) (301) (294) \\
		\hline
		Terminologically Dense Query & Does AMD have a reasonably healthy liquidity profile based on its quick ratio for FY22? If the quick ratio is not relevant to measure liquidity, please state that and explain why? & Current liabilities Accounts payable \$ 2493 \$ 1,321 Accumulated deficit (131) (1451) Accumulated other comprehensive loss (41) (3) Total stockholders equity 54,750 7,497 Common stock, par value \$ 0.01 shares authorized 2250, shares issued 1645 and 1232 \\
		\hline
		Time-Specific Information Request & Has CVS Health reported any materially important ongoing legal battles from 2022, 2021 and 2020 & The Company is named as a defendant in a number of lawsuits The Company is facing multiple lawsuits, including by state Attorneys General, governmental subdivisions and several putative class actions, regarding drug pricing and its rebate arrangements with drug manufacturers. \\
		\hline
		Label noise & Assume that you are a public equities analyst. Answer the following question by primarily using information that is shown in the balance sheet: what is the year end FY2018 net PPNE for 3M? Answer in USD billions. &  Notes to Consolidated Financial Statements are an integral part of this statement. \\
		\hline
	\end{tabular}
\end{table*}

\begin{figure*}[htbp]
    \centering
    \begin{subfigure}{0.48\textwidth}
        \centering
        \includegraphics[width=\textwidth]{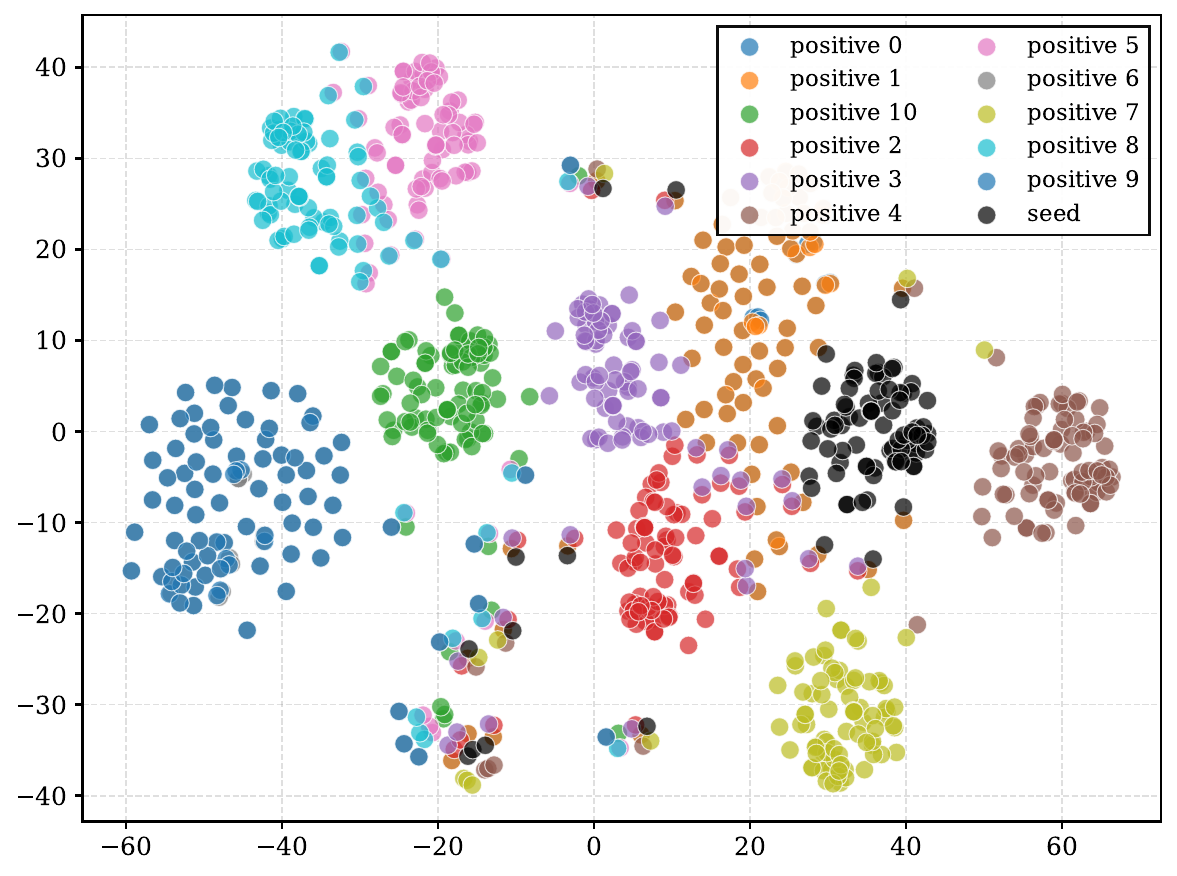}
        \caption{Q1: ``What is the regulatory body that regulates the company?''\\}
        \label{fig:embedding_q1}
    \end{subfigure}
    \hfill
    \begin{subfigure}{0.48\textwidth}
        \centering
        \includegraphics[width=\textwidth]{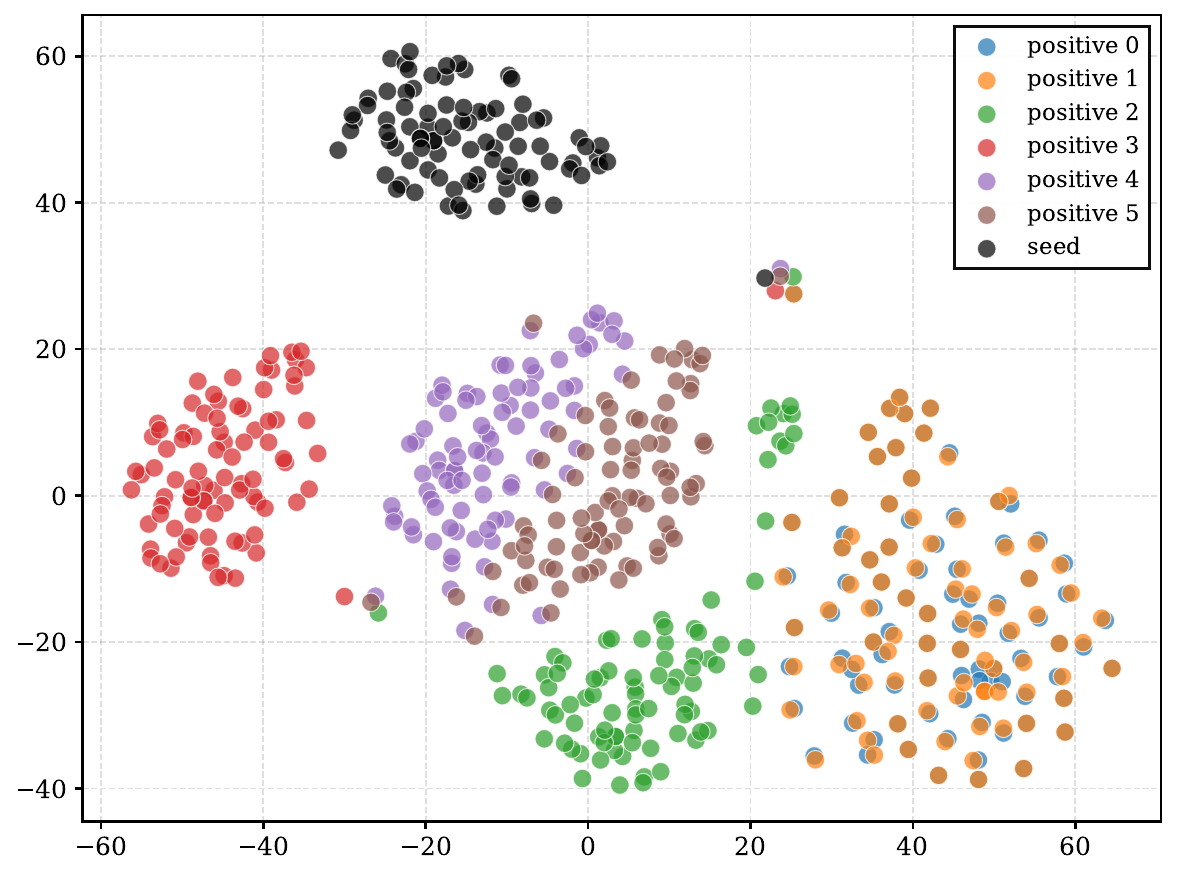}
        \caption{Q2: ``What is the company's current research focus in terms of product development?''}
        \label{fig:embedding_q2}
    \end{subfigure}
    \caption{Plots of $\mathrm{t\mbox{-}SNE}\!\left(\mathcal{S}(q)\right)$,
		the low-dimensional projections of difference vectors
		of positive and negative chunks, with 
		$\mathrm{t\mbox{-}SNE}\!\left(\mathcal{S}_{\mathrm{seed}}(q)\right)$ in black,
		highlighting the benefits of positive example mining in obtaining a more diverse training set.}
    \label{fig:embedding_plots}
\end{figure*}

\subsection{Qualitative Results}
\label{subsec:qualitative}
The first iteration yields the most gain.  On in-domain filing types, the second iteration 
shows smaller effects and mixed results by filing type.
On FinanceBench (OOD) $\mu$NDCG nudges up while MRR is mixed.
To understand why, we analyzed ``promoted passages''---query-chunk pairs where
bi-enc$(1)$ successfully retrieved relevant passages in the top position that bi-enc$(0)$ missed in the top 5--

Examining these improvements across filing classes, we identified several main categories
of promoted passages, including these three:

\begin{itemize}
\item \textbf{Table Retrieval} bi-enc$(1)$ shows enhanced ability to retrieve business metrics from tables,
likely reflecting the higher frequency of tables in SEC filings versus general text used as training data for bi-enc$(0)$.
\item \textbf{Technical Exact Match} bi-enc$(1)$ learned to appropriately weight domain-specific keyphrases (e.g. ``par value'')
in financial contexts.
\item \textbf{Semantic Understanding}  bi-enc$(1)$ recognized conceptual relationships without lexical overlap---connecting
``raising capital'' (in $q$) with ``notes payable...lines of credit'' (in $c$)
or ``risks that could impact the company's business'' (in $q$) with factors ``adversely affecting capital and credit markets'' (in $c$).
\end{itemize}

For the three classes (10-Q, 8-K, ECT) in the FinanceBench dataset for which bi-enc$(1)$ had improved aggregate metrics over bi-enc$(0)$,
we observed similar categories of promoted passages.  For the final class, 10-K, we attempted to diagnose the lack
of improvement by examining the inverse phenomenon of ``demoted (relevant) passages''.  We thereby observed several
patterns prevalent in the (human) SME-written queries of FinanceBench which are likely not adequately represented
in our training data, for example: compound queries asking for several things at once; queries mentioning accounting
concepts such as ``quick ratio'' which need to be unpacked, using external or domain knowledge, into multiple queries
for proper retrieval; queries containing ``distractor'' terms which degrade precision.   
We will propose methods of addressing this gap in Section \ref{sec:discussion}, below.  

In the context of the in-domain SEC filings corpus we define a ``promoted passage'' (with respect to query $q$)
as a $c\in C(d)$ such that 
\begin{enumerate}
	\item $r_{q,c}=4$
	\item $c\in C_d(q,0,1) - C_d(q,5,0)$,
\end{enumerate}
that is $c$ is (within $d$) 
the closest passage to $q$ according to bi-enc$(1)$, but would not be retrieved within the top $5$
under bi-enc$(0)$.  One major pattern seen when the queries asks for relatively straightforward
business metrics, the promoted passage is often a (section of a) table, indicating that the fine-tuned
model does a better job of recognizing when to retrieve tables than the baseline model.  Another
common pattern that is seen is that when a technical financial reporting term (example ``\textit{par}'' as in 
``\textit{par} value'') has an exact match in the ``promoted passage''.  This leads us to speculate
that the fine-tuning causes the model to act more like an exact-keyword match in certain respects, that is to 
weigh more highly exact matches of such technical terms.  Conversely, the fine-tuned
model on occasion appears to exhibit greater semantic understanding of the query than the out-of-box model, as in the case of this example:
``What steps is the company taking to mitigate the risk of not being able to raise the necessary capital to execute its business strategy?''.
One promoted passage for this query is ``...Management plans to address the concerns, as needed, by (a) utilizing recent financing obtained through notes payable; (b) utilizing current lines of credit.''  Although the concerns are not explicitly identified within 
the passage as being about being able to raise capital, an expert reader, and also the model 
are able to infer from the nature of the plans to address the concerns that these concerns are in fact related to raising capital. More examples can be found in Table~\ref{tab:example-passages-filing-corpus} 
for the filings corpus and FinanceBench, respectively.

Before explaining ``promoted passage'' and ``demoted passage'' for FinanceBench we explain how we propagated labels from the FinanceBench dataset to our chunks.
FinanceBench is mainly a Question Answering dataset rather than a relevance dataset, but it has certain passages (not necessarily lining up in any straightforward way with our chunks, and of very different lengths) marked as ``evidence'' chunks for the answer.  Up to a first approximation, we consider a chunk in our chunking over the FinanceBench corpus
(the documents which occur as documents in the evaluation set) relevant if and only if the chunk intersects at least one evidence chunk (considered as a character span
in the original document).  There is one major adjustment to this simple picture: if an intersection is very ``minor'', then the chunk is not considered relevant.
To define what we mean by ``minor'', denote the chunk in question by $c$ and the evidence passage by $p$.  Then (by definition) the intersection is ``minor''
if $\mathrm{len}(c\cap p)\leq\frac{1}{3}\min(\mathrm{len}(c), \mathrm{len}(p))$.  Note that this can occur only if $c$ is the first or last chunk to intersect $p$,
because an ``interior'' chunk of the intersection will intersect along its entire length. This was done to decrease the label noise, or false positives,
in which a chunk had a very short and insignificant overlap with the evidence passage, but was truly irrelevant to the query.  In spite of this provision,
there was still a certain amount of label noise, according to the manual inspection of the results, as shown in the table below, and this may have
an impact on the reported aggregate metrics considering the modest size of the dataset.

Because \textit{every} chunk in FinanceBench is labeled with a relevance, unlike in the case of our filings dataset, where only a sample
(including the top $k$ ranks under the models) is labeled, we are able to 
calculate \textit{overall} NDCG and MRR, not just versions of these metrics \texttt{@}$k$.  For the same reason,
in the context of the Financebench dataset,
we can use a more natural and symmetrical definition of \textit{promoted}
and \textit{demoted} passages. 
Namely, let $q$ and $c\in C(d)$ be fixed with $r_{q,c}=4$, so that $c$ is relevant to $q$.
Denote by $\rho_j(c)$ the rank of $c\in C(d)$ under bi-enc$(j)$.  Consider
\[
	\Delta_i(c) := \rho_{i}(c) -\rho_{i+1}(c).
	\] 
Noting that ranks are numbered in an increasing fashion from closest to furthest from $q$,
we define a promoted (resp. demoted) passage $c$ as one for which $\Delta_i(c)>\mu$ (resp. $<\mu$), $\mu>0$
 some margin. After sorting all of $C(D_{\rm test})$ by the values of $\Delta_i(c)$ in descending order,
 we define the ``most promoted''
(resp. ``most demoted'') passages for a corpus as the head $C(D_{\rm test})[:50]$
(resp. tail $C(D_{\rm test})[-50:]$), and a selection of these is what we actually present in Tables \ref{tab:example-passages-filing-corpus}
and \ref{tab:example-passages-financebench-negatives}.

\section{Analysis of Positive Example Mining Benefits}
Intuitively, the bi-encoder model learns by contrasting embeddings of passages judged relevant 
versus irrelevant for the same query, effectively focusing on difference vectors between these pairs. 
The more diverse the empirical distribution of such differences we expose during training, the more robustly it can internalize the semantics of the query. 
When relying only on an inPars-style sampling strategy (as in  \citet{anderson2024greenback}), 
the training signal is restricted to a narrow subset anchored on a single seed positive, which we visualize as black points. 
Augmenting training with mined positives from our method introduces many additional, complementary difference directions (colored points), 
broadening coverage across the space. In low-dimensional visualizations, the inPars-only subset typically occupies approximately a single cluster, 
whereas the augmented set spans multiple clusters; collectively, this expanded, multi-modal support helps explain the superior performance of our approach.

Each plotted training point is a difference vector:
$$
\mathbf{d}\!\left(q;\, c_{\mathrm{rel}}, c_{\mathrm{irrel}}\right) \;=\; \mathbf{c}_{\mathrm{rel}} \;-\; \mathbf{c}_{\mathrm{irrel}}.
$$

In Figure~\ref{fig:embedding_plots} we visualize all differences where the positive is highly relevant and the negative is less relevant:
$$
\mathcal{S}(q) \;=\; \left\{\, \mathbf{c}_{\mathrm{rel}} \;-\; \mathbf{c}_{\mathrm{irrel}} \;\middle|\; r\!\left(q, c_{\mathrm{rel}}\right)=4,\; r\!\left(q, c_{\mathrm{irrel}}\right)<3 \right\}.
$$
The inPars-only subset anchored on the seed positive (plotted in black):
$$
\mathcal{S}_{\mathrm{seed}}(q) \;=\; \left\{\, \mathbf{c}_{\mathrm{rel}} \;-\; \mathbf{c}_{\mathrm{irrel}} \in \mathcal{S}(q) \;\middle|\; c_{\mathrm{rel}} = c_{\mathrm{seed}} \right\}.
$$
For visualization we use a two-dimensional t-SNE embedding; black points denote the inPars-only subset and colored points are additional mined positives:
$$
\mathrm{t\mbox{-}SNE}\!\left(\mathcal{S}(q)\right).
$$
A commonly used metric in literature, t-SNE~\cite{maaten2008visualizing} is a dimensionality reduction technique 
that is particularly well-suited for visualizing high-dimensional data, as it preserves local structures when projecting to lower-dimensional spaces.
For the plots, we select two exemplary queries: (a) 
"What is the regulatory body that regulates the company?" (Figure \ref{fig:embedding_q1})
 and (b) "What is the company's current research focus in terms of product development?" (Figure \ref{fig:embedding_q2}),
 both associated to the 10-K document class.

\section{Conclusions, Future Work}
\label{sec:discussion}
We showed that an iterative pipeline using mining of positive and negative examples from a large corpus helps
gather training data for domain adaptation of a retrieval model, though most of the benefits to aggregate metrics come from 
a single iteration.  The main limitations of the study are first that passage identification via LLM relies on the current best model
with sampling heuristics, missing many important aspects. Second, that the only public
benchmark (as of time of writing) in this area, namely FinanceBench, while of high quality, is relatively small.  
This demonstrates that query expansion methods are needed either using knowledge graph methods, e.g., ``GraphRAG'' \cite{edge2024graphRAG}, or integrating external knowledge bases.  Alternatively, ``agentic'' models with reasoning capabilities could create more specific queries better suited as anchors in bi-encoder based retrieval.
The agentic model could also examine retrieval results to plan better candidate retrieval from the corpus through
tool use, query augmentation, and improved heuristics.  Rather than capping the investigation into domain
adaptation of retrieval models, we believe this work establishes a solid foundation for broader exploration in the indicated directions.

\balance
\IEEEtriggeratref{5}
\bibliographystyle{plainnat}
\bibliography{bibrepo}

@article{tang2024we,
  title={Do We Need Domain-Specific Embedding Models? An Empirical Investigation},
  author={Tang, Yixuan and Yang, Yi},
  journal={arXiv preprint arXiv:2409.18511},
  year={2024}
}

@inproceedings{li2024retrieval,
  title={Retrieval augmented generation or long-context llms? a comprehensive study and hybrid approach},
  author={Li, Zhuowan and Li, Cheng and Zhang, Mingyang and Mei, Qiaozhu and Bendersky, Michael},
  booktitle={Proceedings of the 2024 Conference on Empirical Methods in Natural Language Processing: Industry Track},
  pages={881--893},
  year={2024}
}

@inproceedings{bonifacio2022inpars,
  title={Inpars: Unsupervised dataset generation for information retrieval},
  author={Bonifacio, Luiz and Abonizio, Hugo and Fadaee, Marzieh and Nogueira, Rodrigo},
  booktitle={Proceedings of the 45th International ACM SIGIR Conference on Research and Development in Information Retrieval},
  pages={2387--2392},
  year={2022}
}

@misc{sec_edgar,
  author       = {{U.S. Securities and Exchange Commission}},
  title        = {EDGAR - Electronic Data Gathering, Analysis, and Retrieval system},
  year         = 2025,
  url          = {https://www.sec.gov/edgar.shtml},
}

@article{li2023towards,
	title={Towards general text embeddings with multi-stage contrastive learning},
	author={Li, Zehan and Zhang, Xin and Zhang, Yanzhao and Long, Dingkun and Xie, Pengjun and Zhang, Meishan},
	journal={arXiv preprint arXiv:2308.03281},
	year={2023}
}

@article{wang2020minilm,
	title={Minilm: Deep self-attention distillation for task-agnostic compression of pre-trained transformers},
	author={Wang, Wenhui and Wei, Furu and Dong, Li and Bao, Hangbo and Yang, Nan and Zhou, Ming},
	journal={Advances in Neural Information Processing Systems},
	volume={33},
	pages={5776--5788},
	year={2020}
}

@misc{artificialanalysis2024,
	title = {Artificial Analysis AI Review 2024 Highlights},
	author = {Artificial Analysis},
	howpublished = {\url{https://artificialanalysis.ai/downloads/ai-review/2024/Artificial-Analysis-AI-Review-2024-Highlights.pdf}},
	year = {2024},
	note = {Retrieved January 15, 2025},
}

@article{tang2025finmteb,
  title={FinMTEB: Finance Massive Text Embedding Benchmark},
  author={Tang, Yixuan and Yang, Yi},
  journal={arXiv preprint arXiv:2502.10990},
  year={2025}
}

@inproceedings{anderson2024greenback,
    title = "Greenback Bears and Fiscal Hawks: Finance is a Jungle and Text Embeddings Must Adapt",
    author = "Anderson, Peter  and
      Janardhanan, Mano Vikash  and
      He, Jason  and
      Cheng, Wei  and
      Flanagan, Charlie",
    editor = "Dernoncourt, Franck  and
      Preo{\c{t}}iuc-Pietro, Daniel  and
      Shimorina, Anastasia",
    booktitle = "Proceedings of the 2024 Conference on Empirical Methods in Natural Language Processing: Industry Track",
    month = nov,
    year = "2024",
    address = "Miami, Florida, US",
    publisher = "Association for Computational Linguistics",
    pages = "362--370"
}

@inproceedings{peng2021domain,
  title={Is domain adaptation worth your investment? comparing BERT and FinBERT on financial tasks},
  author={Peng, Bo and Chersoni, Emmanuele and Hsu, Yu-Yin and Huang, Chu-Ren and others},
  year={2021},
  booktitle={Proceedings of the Third Workshop on Economics and Natural Language Processing},
  pages={37--44},
  address={Punta Cana, Dominican Republic},
  month={November},
  organization={Association for Computational Linguistics (ACL)},
}

@misc{meta-llama-llama-3.1-70b-instruct,
	title = {Llama-3.1-70B-Instruct Model Card},
	author = {Meta AI},
	howpublished = {\url{https://huggingface.co/meta-llama/Llama-3.1-70B-Instruct}},
	year = {2024},
	note = {Retrieved January 15, 2025},
}

@online{captide_agentic_rag,
  author = {CapTide},
  title = {How to Do Agentic RAG on SEC Edgar Filings},
  year = {2024}, 
  url = {https://www.captide.co/insights/how-to-do-agentic-rag-on-sec-edgar-filings},
  urldate = {2025-03-05}, 
  note = {Accessed March 5, 2025}
}

@inproceedings{Choe2025Hierarchical,
  author    = {Jaeyoung Choe and Jihoon Kim and Woohwan Jung},
  title     = {Hierarchical Retrieval with Evidence Curation for Open-Domain Financial Question Answering on Standardized Documents},
  booktitle = {Findings of the Association for Computational Linguistics: ACL 2025},
  pages     = {16663--16681},
  year      = {2025}
}

@inproceedings{Reddy2024DocFinQA,
  author    = {Varshini Reddy and Rik Koncel-Kedziorski and Viet Dac Lai and Michael Krumdick and Charles Lovering and Chris Tanner},
  title     = {{D}oc{F}in{QA}: A Long-Context Financial Reasoning Dataset},
  booktitle = {Proceedings of the 62nd Annual Meeting of the Association for Computational Linguistics (Volume 2: Short Papers)},
  pages     = {445--458},
  year      = {2024},
  month     = aug,
  address   = {Bangkok, Thailand},
  publisher = {Association for Computational Linguistics},
}

@inproceedings{zhao-etal-2024-optimizing,
    title = "Optimizing {LLM} Based Retrieval Augmented Generation Pipelines in the Financial Domain",
    author = "Zhao, Yiyun and Singh, Prateek and Bhathena, Hanoz and Ramos, Bernardo and Joshi, Aviral and Gadiyaram, Swaroop and Sharma, Saket",
    editor = "Yang, Yi and Davani, Aida and Sil, Avi and Kumar, Anoop",
    booktitle = "Proceedings of the 2024 Conference of the North American Chapter of the Association for Computational Linguistics",
    month = jun,
    year = "2024",
    address = "Mexico City, Mexico",
    publisher = "ACL",
    pages = "279--294"
}

@manual{SECEdgarFilerManualVol2,
  title        = {{EDGAR} Filer Manual Volume {II}: {EDGAR} Filing},
  author       = {{U.S. Securities and Exchange Commission}},
  organization = {{U.S. Securities and Exchange Commission}},
  year         = {2025},
  month        = {sep},
  version      = {76},
  url          = {https://www.sec.gov/edgar/filer-manual},
}

@article{wang2024omnieval,
	title={OmniEval: An Omnidirectional and Automatic RAG Evaluation Benchmark in Financial Domain},
	author={Wang, Shuting and Tan, Jiejun and Dou, Zhicheng and Wen, Ji-Rong},
	journal={arXiv preprint arXiv:2412.13018},
	year={2024}
}

@article{islam2023financebench,
	title={Financebench: A new benchmark for financial question answering},
	author={Islam, Pranab and Kannappan, Anand and Kiela, Douwe and Qian, Rebecca and Scherrer, Nino and Vidgen, Bertie},
	journal={arXiv preprint arXiv:2311.11944},
	year={2023}
}

@book{cohen1988statistical,
  title={Statistical Power Analysis for the Behavioral Sciences},
  author={Cohen, Jacob},
  year={1988},
  edition={2nd},
  publisher={Lawrence Erlbaum Associates},
  address={Hillsdale, NJ},
  pages={567},
  isbn={0-8058-0283-5}
}

@article{edge2024graphRAG,
	author       = {Darren Edge and
	Ha Trinh and
	Newman Cheng and
	Joshua Bradley and
	Alex Chao and
	Apurva Mody and
	Steven Truitt and
	Jonathan Larson},
	title        = {From Local to Global: {A} Graph {RAG} Approach to Query-Focused Summarization},
	journal      = {arXiv preprint arXiv:2404.16130},
	year         = {2024}
}

@article{maaten2008visualizing,
  title={Visualizing data using t-SNE},
  author={Maaten, Laurens van der and Hinton, Geoffrey},
  journal={Journal of machine learning research},
  volume={9},
  number={Nov},
  pages={2579--2605},
  year={2008}
}

\end{document}